# KINEMATIC CALIBRATION OF ORTHOGLIDE-TYPE MECHANISMS

**Anatoly Pashkevich[1,2], Damien Chablat[2], Philippe Wenger[2]**

[1]*Robotic Laboratory, Department of Automatic Control*
*Belarusian State University of Informatics and Radioelectronics*
*6 P.Brovka St., Minsk 220027, Belarus, e-mail: pap@bsuir.unibel.by*

[2]*Institut de Recherche en Communications et Cybernétique de Nantes*
*1, rue de la Noë B.P. 6597, 44321 Nantes Cedex 3, France*
*e-mals: {Damien.Chablat, Philippe.Wenger }@irccyn.ec-nantes.fr*

Abstract: *The paper proposes a novel calibration approach for the Orthoglide-type mechanisms based on observations of the manipulator leg parallelism during motions between the prespecified test postures. It employs a low-cost measuring system composed of standard comparator indicators attached to the universal magnetic stands. They are sequentially used for measuring the deviation of the relevant leg location while the manipulator moves the TCP along the Cartesian axes. Using the measured differences, the developed algorithm estimates the joint offsets that are treated as the most essential parameters to be adjusted. The sensitivity of the measurement methods and the calibration accuracy are also studied. Experimental results are presented that demonstrate validity of the proposed calibration technique.*

## 1. INTRODUCTION

Parallel kinematic machines (PKM) are commonly claimed to offer several advantages over serial manipulators, like high structural rigidity, better payload-to-weight ratio, high dynamic capacities and high accuracy (Merlet, 2000; Tlusty, *et al.*, 1999). As stated by a number of authors (Tsai, 1999; Wenger, *et al.*, 1999), the conventional serial kinematic structures have already achieved their performance limits, which are bounded by high component stiffness required to support sequential joints, links and actuators. Thus, the PKM are prudently considered as promising alternatives to their serial counterparts that offer faster, more flexible, more accurate and less costly solutions.

However, while the PKM usually demonstrate a much better repeatability compared to serial mechanisms, they may not necessarily posses a better accuracy, which is limited by manufacturing/assembling errors in numerous links and passive joints (Wang, and Masory, 1993). Besides, for the non-Cartesian parallel architectures, some kinematic parameters (such as the encoder offsets) cannot be determined by direct measurement. These motivate intensive research on PKM calibration.

Similar to the serial manipulators, the PKM calibration procedures are based on the minimisation of a parameter-dependent error function, which incorporates residuals of the kinematic equations. For the parallel manipulators, the inverse kinematic equa-tions are considered computationally more efficient, since the most of the PKM admits a closed-form solution of the inverse kinematics (Innocenti, 1995; Iurascu, and Park, 2003; Huang, *et al*., 2005). But the main difficulty with the inverse-kinematics-based calibration is the full-pose measurement requirement, which is very hard to implement accurately (Jeong, *et al.*, 2004.

Recently, several hybrid calibration methods were proposed that utilize intrinsic properties of a particular parallel machine allowing to extract the full set of the model parameters (or the most essential of them) from a minimum set of the pose error measurements. An innovative approach for the I4 parallel mechanism has been developed by Renaud *et al* (2004, 2006) who applied vision to perform measurement on manipulator legs, which connect the end-effector to the base-mounted linear actuators.

This paper focuses on the Orthoglide-type mechanisms, which kinematic architecture admits parallel leg motions (along the longitudinal axis). From observations the leg parallelism at prespecified postures, the proposed calibration algorithm estimates the joint offsets that are treated as the most essential parameters to be adjusted. The main advantage of this approach is the simplicity and low cost of the measuring system composed of standard comparator indicators attached to the universal magnetic stands. They are sequentially used for measuring the deviation of the relevant leg location while the manipulator moves the TCP along the Cartesian axes.

The remainder of the paper is organised as follows. Section 2 describes the manipulator geometry, its inverse and direct kinematics, and also contains the sensitivity analysis of the leg parallelism at the examined postures with respect to the joint offsets. Section 3 focuses on the measurements and parameter identification. Section 4 contains experimental results and their discussion, while Section 5 summarises the main results.

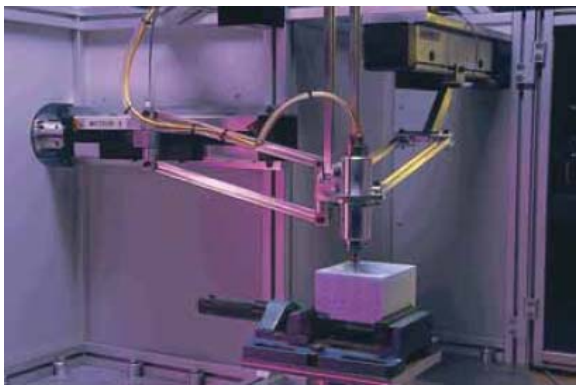

Fig. 1. The Orthoglide mechanism

## 2. KINEMATIC MODEL

### 2.1 Manipulator geometry

The Orthoglide is a three degree-of-freedom mechanism actuated by three linear drives with mutually orthogonal axes (Chablat and Wenger, 2003). Its kinematic architecture is presented in Fig. 1 and includes three identical parallel chains that are formally described as *PRPaR*, where *P*, *R* and *Pa* denote the prismatic, revolute, and parallelogram joints.

The mechanism input is made up by three actuated orthogonal prismatic joints. The output machinery (with a tool mounting flange) is connected to the prismatic joints through a set of three kinematic chains that further referred as manipulator "legs". The legs are made with articulated parallelograms that are oriented in a manner that the output body is restricted to the translational movements only.

A specific feature of the Orthoglide mechanism, which will be further used for the calibration, is displayed during the end-effector motions along the Cartesian axes. For example, for the *x*-axis motion in the Cartesian space, the sides of the *x*-leg parallelogram must also retain strictly parallel to the *x*-axis. Hence, the observed deviation of the mentioned parallelism may be used as the data source for calibration algorithms.

### 2.2 Modelling assumptions

Following previous studies on the parallel mechanism accuracy (Wang, and Massory, 1995, Renaud, *et al.*, 2004; Caro, *et al.*, 2006), the influence of the joint/link defects is assumed negligible compared to the joint positioning errors mainly caused by the encoder offsets. This validates the following modelling assumptions:

(i) the manipulator parts are supposed to be rigid-bodies connected by perfect joints;
(ii) the articulated parallelograms are assumed to be identical and perfect, which insure that their sides stay parallel in pares for any motions;
(iii) the manipulator legs (composed of one prismatic joint, one parallelogram, and three revolute joints) are identical and generate a five-DOF motion each;
(iv) the linear actuator axes are considered mutually orthogonal and intersected in a single point to insure a translational three-DOF movement of the end-effector.

The actuator encoders are assumed to be perfect but their location (zero position) is defined with some errors that are treated as the offsets to be estimated. Using these assumptions, there will be developed a convenient calibration methodology based on the observation of the parallel gel motions.

## 2.3 Basic equations

Under the above assumptions, the Orthoglide can be presented by a simplified model, which consists of three rod-links connected by spherical joints to the tool centre point (TCP) at one side and to the corresponding prismatic joints at another side. Thus, if the origin of a reference frame is located at the intersection of the prismatic joint axes and the $x$, $y$, $z$-axes are directed along them, the manipulator geometry may be described by the equations

$$[p_x - (\rho_x + \Delta\rho)]^2 + p_y^2 + p_z^2 = L^2$$
$$p_x^2 + [p_y - (\rho_y + \Delta\rho_y)]^2 + p_z^2 = L^2 \quad (1)$$
$$p_x^2 + p_y^2 + [p_z - (\rho_z + \Delta\rho_z)]^2 = L^2$$

where $\mathbf{p} = (p_x, p_y, p_z)$ is the output position vector, $\mathbf{\rho} = (\rho_x, \rho_y, \rho_z)$ is the input vector of the joints variables, $\Delta\mathbf{\rho} = (\Delta\rho_x, \Delta\rho_y, \Delta\rho_z)$ is the encoder offset vector, and $L$ is the length of the parallelogram principal links. Hence, within the adopted model, four parameters ($\Delta\rho_x$, $\Delta\rho_y$, $\Delta\rho_z$, $L$) define the manipulator geometry, but because of the rather tough manufacturing tolerances, the leg length is assumed to be known and only the joint offsets ($\Delta\rho_x$, $\Delta\rho_y$, $\Delta\rho_z$) are in the focus of the proposed calibration technique.

It should be noted that for the nominal "mechanical-zero" posture corresponding to the joints variables $\mathbf{\rho}_0 = (L, L, L)$, the $x$-, $y$- and $z$-legs are oriented strictly parallel to the relevant Cartesian axes. But the joint offsets cause the deviation of the "zero" location and corresponding deviation of the leg parallelism with respect to the manipulator base surface. The latter may be (i) measured and (ii) computed applying the direct kinematic algorithm for the joint variables $\mathbf{\rho} = (L+\Delta\rho_x, L+\Delta\rho_y, L+\Delta\rho_z)$. However, the capability of this simple technique is limited by evaluating the offset of the $z$-axis encoder only, since the Orthoglide mechanical design does not allow making similar measurements for the remaining pairs of the legs, with respect to the $xz$- and $yz$-planes.

## 2.4 Inverse and direct kinematics

To derive calibration equations, first let us expand some previous results on the Orthoglide kinematics (Pashkevich *et al.*, 2005) taking into account the encoder offsets. The *inverse kinematic* relations are derived from the equations (1) in a straightforward way and only slightly differ from the "nominal" case

$$\rho_x = p_x + s_x\sqrt{L^2 - p_y^2 - p_z^2} - \Delta\rho_x$$
$$\rho_y = p_y + s_y\sqrt{L^2 - p_x^2 - p_z^2} - \Delta\rho_y \quad (2)$$
$$\rho_z = p_z + s_z\sqrt{L^2 - p_x^2 - p_y^2} - \Delta\rho_z$$

where $s_x$, $s_y$, $s_z \in \{\pm 1\}$ are the configuration indices defined for the "nominal" manipulator as signs of $(\rho_x - p_x)$, $(\rho_y - p_y)$, and $(\rho_z - p_z)$ respectively. It is obvious that expressions (2) define eight different solutions to the inverse kinematics, however the Orthoglide assembling and joint limits reduce this set for a single case corresponding to the $s_x = s_y = s_z = 1$.

For the *direct kinematics*, the equations (1) can be subtracted pair-to-pair that gives the following expression for the unknowns $p_x$, $p_y$, $p_z$

$$p_i = \frac{\rho_i + \Delta\rho_i}{2} + \frac{t}{\rho_i + \Delta\rho_i}; \quad i \in \{x, y, z\} \quad (3)$$

where $t$ is an auxiliary scalar variable. This reduces the direct kinematics to the solution of a quadratic equation $at^2 + bt + c = 0$ with the coefficients

$$a = \prod_{i \neq j}(\rho_i + \Delta\rho_i)(\rho_j + \Delta\rho_j); \quad b = \prod_i(\rho_i + \Delta\rho_i)^2;$$
$$c = \sum_i(\rho_i + \Delta\rho_i)^2/4 - L^2; \quad i, j \in \{x, y, z\}$$

From two possible solutions that give the quadratic formula, the Orthoglide prototype (see Fig. 1) admits only one expressing as $t = (-b + m\sqrt{b^2 - 4ac})/2a$.

## 2.5 Sensitivity analysis

To evaluate the encoder offset influence on the legs parallelism with respect to the Cartesian planes *XY*, *YZ*, and *YZ*, let us derive the differential relations for the TCP deviation for three types of the Orthoglide postures: *mechanical zero*" posture and "*maximum/minimum displacement*" postures for the directions $x$, $y$, $z$. These postures are of particular interest for the calibration since in the "nominal" case (no encoder offsets) at least one leg is parallel to the corresponding pair of the Cartesian planes.

The desired differential relations may be derived from the Orthoglide Jacobian, the inverse of which is obtained from (1) as

$$\mathbf{J}^{-1}(\mathbf{p}, \mathbf{\rho}) = \begin{bmatrix} 1 & \dfrac{p_y}{p_x - \rho_x} & \dfrac{p_z}{p_x - \rho_x} \\ \dfrac{p_x}{p_y - \rho_y} & 1 & \dfrac{p_z}{p_y - \rho_y} \\ \dfrac{p_x}{p_z - \rho_z} & \dfrac{p_y}{p_z - \rho_z} & 1 \end{bmatrix} \quad (4)$$

For the "*mechanical zero*" posture, the differential relations are derived in the neighbourhood of the point $\mathbf{p}_0 = (0, 0, 0)$ and $\mathbf{\rho}_0 = (L, L, L)$, which gives the identity Jacobian matrix $\mathbf{J}(\mathbf{p}_0, \mathbf{\rho}_0) = \mathbf{I}_{3\times 3}$. Hence, in this case the TCP displacement is related to the joint offsets by trivial equations $\Delta p_i = \Delta\rho_i; \quad i \in \{x, y, z\}$. Taking into the account the Orthoglide geometry, this deviation may be estimated by evaluating parallelism of the legs with respect to the Cartesian planes. How-ever, as mentioned in sub-section 2.3, this technique is feasible for *z*-direction only and may produce estimation of $\Delta\rho_z$ merely.

For the *"maximum displacement"* posture in the *x*-direction, the differential relations are derived in the neighbourhood of the point $\mathbf{p}=(L\sin\alpha, 0, 0)$ and $\mathbf{\rho}=(L+LS_\alpha, LC_\alpha, LC_\alpha)$, where $\alpha$ is the angle between the *y*-, *z*-legs and corresponding Cartesian axes, and $S_\alpha = \sin\alpha$; $C_\alpha = \cos\alpha$. This gives the inverse Jacobian as a lower triangle matrix

$$\mathbf{J}(\mathbf{p},\mathbf{\rho}) = \begin{bmatrix} 1 & 0 & 0 \\ T_\alpha & 1 & 0 \\ T_\alpha & 0 & 1 \end{bmatrix} \quad (5)$$

where $T_\alpha = \tan\alpha$. Hence, the differential equations for the TCP displacement may be written as

$$\begin{aligned} \Delta p_x &= \Delta\rho_x; \\ \Delta p_y &= T_\alpha \Delta\rho_x + \Delta\rho_y; \\ \Delta p_z &= T_\alpha \Delta\rho_x + \Delta\rho_z; \end{aligned} \quad (6)$$

and the joint offset influences on the TCP deviation is estimated by factors 1.0 and $T_\alpha$. It is also worth mentioning that measurement of the *x*-leg parallelism with respect to the *XY*-plane gives an equation for estimating the offset $\Delta\rho_x$ (provided that the offset $\Delta\rho_z$ has been obtained from the "mechanical zero").

Similar results are valid for the *"maximum displacement"* postures in the *y*- and *z*-directions (differing by the indices only), and also for the *"minimum displacement"* postures. In the latter case, the angle $\alpha$ should be computed as $\alpha = \operatorname{asin}(\rho_{\min}/L)$. Hence, the leg parallelism is rather sensitive to the joint offsets and relevant deviations $\Delta p_x$, $\Delta p_y$, $\Delta p_z$, and may be used for the offset identification.

## 3. CALIBRATION METHODOLOGY

### 3.1 Measurement technique

To identify the model parameters, we propose a single-sensor measurement techniques for the leg/surface parallelism (Fig. 2). It is based on fixed location of the measuring device for two distinct leg postures corresponding to the minimum/maximum of the joint coordinates. Relevant calibration experiment consists of the following steps:

**Step 1**. Move the manipulator to the *"mechanical zero"*; locate two gauges in the middle of the X-leg (parallel to the axes Y and Z); get their readings.

**Step 2**. Move the manipulator to the *"X-maximum"* and *"X-minimum"* postures, get the gauge readings, and compute differences $\Delta y_x^+, \Delta z_x^+, \Delta y_x^-, \Delta z_x^-$.

**Step 3+**. Repeat steps 1, 2 for the Y- and Z-legs and compute corresponding differences.

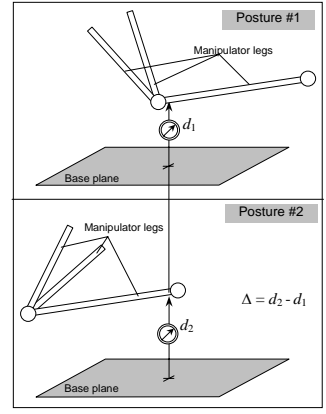

Fig. 2. Measuring the leg/surface parallelism

### 3.2 Calibration equations

The calibration equations can be derived using expressions from sub-section 2.5. First, it is required to define the gauge initial location. For the X-leg, it is the midpoint of the line segment bounded by the TCP ($\Delta\rho_x$, $\Delta\rho_y$, $\Delta\rho_z$) and the centre of the X- joint ($L+\Delta\rho_x$, 0, 0). This yields the following expression for the *X*-leg midpoint: $(L/2 + \Delta\rho_x;\ \Delta\rho_y/2;\ \Delta\rho_z/2)$. For the *"X-maximum"* posture, the X-leg location is also defined by same two points, but their coordinates are $(LS_\alpha + \Delta\rho_x; T_\alpha\Delta\rho_x + \Delta\rho_y; T_\alpha\Delta\rho_x + \Delta\rho_z)$ and $(L+LS_\alpha + \Delta\rho_x;\ 0;\ 0)$ respectively. Then, it may be written the equations of a straight-line passing along the X-leg and computed the point corresponding to $x = L/2 + \Delta\rho_x$. Hence, finally, the deviations of the X-leg measurements may be expressed as

$$\begin{aligned} \Delta y_x^+ &= (0.5 + S_\alpha)T_\alpha\ \Delta\rho_x + S_\alpha\ \Delta\rho_y; \\ \Delta z_x^+ &= (0.5 + S_\alpha)T_\alpha\ \Delta\rho_x + S_\alpha\ \Delta\rho_z \end{aligned} \quad (7)$$

Similar approach may be applied to the "X-minimum" posture, as well as to the equivalent postures for the *Y*- and *Z*-legs. This gives the system of twelve linear equations in three unknowns

$$\begin{bmatrix} b_1 & c_1 & 0 \\ c_1 & b_1 & 0 \\ b_2 & c_2 & 0 \\ c_2 & b_2 & 0 \\ \hline 0 & b_1 & c_1 \\ 0 & c_1 & b_1 \\ 0 & b_2 & c_2 \\ 0 & c_2 & b_2 \\ \hline b_1 & 0 & c_1 \\ c_1 & 0 & b_1 \\ b_2 & 0 & c_2 \\ c_2 & 0 & b_2 \end{bmatrix} \cdot \begin{bmatrix} \Delta\rho_x \\ \Delta\rho_y \\ \Delta\rho_z \end{bmatrix} = \begin{bmatrix} \Delta x_y^+ \\ \Delta y_x^+ \\ \Delta x_y^- \\ \Delta y_x^- \\ \Delta y_z^+ \\ \Delta z_y^+ \\ \Delta y_z^- \\ \Delta z_y^- \\ \Delta x_z^+ \\ \Delta z_x^+ \\ \Delta x_z^- \\ \Delta z_x^- \end{bmatrix} \quad (8)$$

where $b_i = \sin\alpha_i$; $c_i = (0.5 + \sin\alpha_i)\tan\alpha_i$ and the angles $\alpha_1, \alpha_2$ are computed for the maximum and minimum postures respectively. The reduced version of this system may be obtained if to assess the leg/plane parallelism by the difference between the "maximum" and "minimum" postures. The latter leads to the system of six linear equations in three unknowns.

## 4. EXPERIMENTAL RESULTS

### 3.1 Experimental setup

The measuring system is composed of standard comparator indicators attached to the universal magnetic stands allowing fixing them on the manipulator bases. The indicators have resolution of 10 μm and are sequentially used for measuring the X-, Y-, and Z-leg parallelism while the manipulator moves between the Max, Min and Zero postures. For each measurement, the indicators are located on the mechanism base in such manner that a corresponding leg is admissible for the gauge contact for all intermediate posters (Fig. 3).

For each leg, the measurements were repeated three times for the following sequence of motions: Zero → Max → Min → Zero→ …. Then, the results were averaged and used for the parameter identification. It should be noted that measurements demonstrated very high repeatability (about 0.02 mm).

### 4.2 Calibration results

The first calibration experiment produced rather high parallelism deviation, up to 2.37 mm, which in the frames of the adopted kinematic model was expected to be reduced down to 1.07 mm only (Table 1). The corresponding residual norm was also rather high and yielded unrealistic estimate of the measurement noise parameter $\sigma \approx 1.0$ mm. It impels to conclude that the mechanism mechanics requires more careful tuning. Consequently, the location of the joint axes was adjusted mechanically to ensure the leg parallelism for the Zero posture.

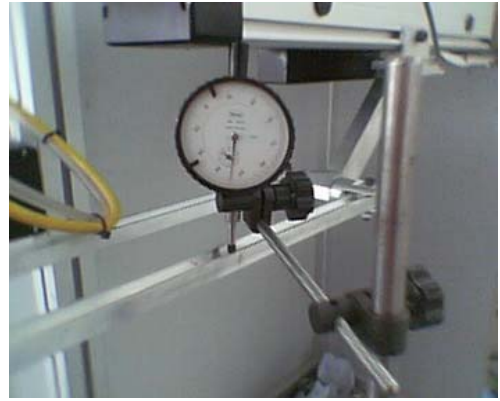

Fig. 3. Experimental Setup

The second calibration experiment (after mechanical tuning) yielded lower parallelism deviations, less than 0.70 mm. For these data, the developed calibration algorithm yielded the joint offsets that are expected to reduce the deviation down to 0.28 mm. Besides, the estimated value of $\sigma \approx 0.28$ mm is rather realistic taking into account both the measurement accuracy and the manufacturing/assembling tolerances. Accordingly, the identified vales of the joint offsets were input into the control software.

The third experiment (validation of the identified parameters) demonstrated good agreement with the expected results. In particular, the maximum parallelism deviation reduced down to 0.34 mm (expected 0.28 mm), while its root-mean-square decreased down to 0.21 mm (expected 0.20 mm). On the other hand, further adjusting of the kinematic model to the new experimental data gives both negligible improvement of the r.m.s. and rather small alteration of the model parameters. It is evident that further reduction of the parallelism deviation is bounded by the manufacturing and assembling errors.

Hence, the calibration results confirm validity of the proposed identification technique and its ability to tune the joint offsets from observations of the leg parallelism.

Table 1. Experimental data and expected improvements of accuracy (model-based)

| Data Source | $\Delta x_y$ mm | $\Delta x_z$ mm | $\Delta y_x$ mm | $\Delta y_z$ mm | $\Delta z_x$ mm | $\Delta z_y$ mm | r.m.s. mm |
|---|---|---|---|---|---|---|---|
| *Initial settings (before mechanical tuning and calibration)* | | | | | | | |
| Experiment #1 | +0.52 | +1.58 | +2.37 | -0.25 | -0.57 | -0.04 | 1.19 |
| Expected improvement | -0.94 | +0.63 | +1.07 | -0.84 | -0.27 | +0.35 | 0.74 |
| *After mechanical tuning (before calibration)* | | | | | | | |
| Experiment #2 | -0.43 | -0.37 | +0.42 | -0.18 | -1.14 | -0.70 | 0.62 |
| Expected improvement | -0.28 | +0.25 | +0.21 | -0.14 | -0.13 | +0.09 | 0.20 |
| *After calibration* | | | | | | | |
| Experiment #3 | -0.23 | +0.27 | +0.34 | -0.10 | -0.09 | +0.11 | 0.21 |
| Expected improvement | -0.29 | +0.23 | +0.25 | -0.17 | -0.10 | +0.08 | 0.20 |

## 5. CONCLUSIONS

This paper proposes a novel calibration approach for the parallel manipulators, which is based on observations of manipulator leg parallelism with respect to some predefined planes. Presented for the Orthoglide-type mechanisms, this approach may be also applied to other manipulator architectures that admit parallel leg motions along the longitudinal axis.

The proposed calibration technique employs a simple and low-cost measuring system composed of standard comparator indicators attached to the universal magnetic stands. They are sequentially used for measuring the deviation of the relevant leg location while the manipulator moves the TCP along the Cartesian axes. From the measured differences, the calibration algorithm estimates the joint offsets that are treated as the most essential parameters to be tuned. The validity of the proposed approach and efficiency of the developed numerical algorithm were confirmed by the calibration experiments with the Orthoglide prototype, which allowed reducing the residual r.m.s. by three times.

To increase the calibration precision, future work will focus on the development of the specific assembling fixture ensuring proper location of the linear actuators and also on the essential theoretical issues such as calibration experiment planning, expanding the set of the identified model parameters and their identifiably analysis, and compensation of the non-geometric errors non detected within the frames of the adopted kinematic model.